%% file: iclr2025_conference.tex
\documentclass{article} % For LaTeX2e
\usepackage{iclr2025_conference,times}

\input{math_commands.tex}

\usepackage{booktabs}
\usepackage{multirow}
\usepackage{hyperref}
\usepackage{url}
\usepackage{graphicx}

\title{A Recipe for Improving Remote Sensing VLM Zero Shot Generalization}
\author{Aviad Barzilai\footnotemark[1]\thanks{Equal Contributors. Correspondence to: George Leifman gleifman@google.com.}, Yotam Gigi\footnotemark[1], 
Amr Helmy\footnotemark[1],Vered Silverman\footnotemark[1], Yehonathan Refael\footnotemark[1], Bolous Jaber,\\ \textbf{Tomer Shekel, George Leifman, Genady Beryozkin} \\
Google Research
}
\usepackage{enumitem, kantlipsum}
\usepackage{multirow}
\usepackage{amsmath}
\usepackage{graphicx}

\iclrfinalcopy % Uncomment for camera-ready version, but NOT for submission.
\begin{document}

\maketitle

\begin{abstract}
Foundation models have had a significant impact across various AI applications, enabling use cases that were previously impossible. Contrastive Visual Language Models (VLMs), in particular, have outperformed other techniques in many tasks. 
However, their prevalence in remote sensing (RS) is still limited, due to the scarcity of diverse remote-sensing visual-language datasets.
In this work we introduce two novel image-caption datasets for training of remote sensing foundation models. The first dataset pairs aerial and satellite imagery with captions generated by Gemini using landmarks extracted from Google Maps. The second dataset utilizes public web images and their corresponding alt-text, filtered for the remote sensing domain, resulting in a diverse dataset with greater breadth in image styles and subject matter.
These datasets are used to pre-train the MaMMUT~\citep{kuo2023mammutsimplearchitecturejoint} VLM architecture, resulting in state-of-the-art generalization performance in zero-shot cross-modal retrieval on well-known public benchmarks. 
Finally, we present our ongoing research to distill image-level knowledge gained in the VLM contrastive training procedure to enhance the model's localization ability. 
Specifically, we iteratively generate pseudo-labels for image regions based on the model's attention maps and use these labels for further training.  
To mitigate noisy attention maps and create robust segmentation masks, we introduce a novel attention-pooling mechanism called the Smooth-Attention-Operation.
\end{abstract}

\section{Introduction}

% Plug in and merge with related work:
% Fill the gap approaches - Domain-Specific Adaptations:
% Fine-Tuning on Remote Sensing Data: fine-tunes CLIP's image encoder on curated datasets of satellite imagery paired with captions (e.g  RemoteCLIP)
% Data Augmentation: converting heterogeneous annotations into unified image-caption formats (e.g., Box-to-Caption, Seg-to-Caption - RemoteCLIP
% )
% Cross-Modality Integration: Incorporating geo-tagged ground-level images alongside satellite imagery helps bridge the gap between natural image features and remote sensing data (sat2cap?)
% Self-supervised learning (e.g., masked autoencoders) and multi-modal architectures that integrate diverse data types like spectral bands or temporal sequences (SatMAE?, DINO-MC?)

Foundation models have demonstrated exceptional performance serving as a basis for diverse downstream tasks by training on large-scale datasets. 
Yet, applying them to remote sensing tasks presents unique challenges, as remote sensing data exhibits distinct characteristics, including complex spatial relationships from orbital viewpoints and lower spatial resolution compared to standard ground-level imagery. 
These differences hinder existing models optimized for ground-level perspectives and high-resolution images.
Furthermore, the remote sensing domain faces labeled data scarcity, including limited availability of paired text-image examples. 
The paucity of datasets with rich textual descriptions accompanying remote sensing images limits the ability to build models that comprehend the context and nuances of the visual data. 

Self-supervised learning techniques, such as contrastive spatial pre-training explored in \citep{mai2023csp}, offer a promising avenue to mitigate this issue by leveraging the inherent spatial information within readily available unlabeled remote sensing data. As the study in \cite{bourcier2024learning} demonstrates, leveraging metadata supervision offers a promising avenue to overcome this limitation, learning effective representations even when paired text-image data is scarce.
% Apart from that, the lack of well pair representation alignment, coupled with a large dataset containing sufficient open-set concepts, the model can achieve strong zero-shot abilities.
% This scarcity hinders the training of models reliant on supervised learning and the critical alignment of visual features with semantic concepts. 
These challenges highlight the need for tailored solutions that address the specific nature of remote sensing data. Similarly, a novel VLM, specifically designed for remote sensing is needed.
Such a model will enable more accurate and insightful analysis for a wide range of applications using remote sensing imagery, including open-vocabulary object detection, zero-shot image segmentation, text-to-image generation and editing, and so on.  Recent research like GeoCLIP \citep{NEURIPS2023_1b57aadd} and SatCLIP \citep{klemmer2024} exemplify the growing recognition of this need, exploring CLIP-inspired methods and location embeddings specifically for geospatial understanding. These efforts, while promising, also underscore the ongoing need for more generalized and robust VLMs in the remote sensing domain, capable of handling the unique challenges of this data.

% \textbf{Vision-Language Models (VLMs).}
Recent advances in VLMs have shown substantial benefits across numerous natural images tasks. CLIP~\citep{radford2021learning} was the first to utilize contrastive loss, aligning images with their corresponding textual descriptions in a shared embedding space. 
% GLIP~\citep{Li2021GroundedLP,zhang2022glipv2} incorporated grounding data into the vision-language framework, demonstrating its applicability in various object-level recognition tasks. 
BLIP~\citep{li2022blip} introduced a bootstrapping mechanism to refine noisy captions, further enhancing performance in vision-language tasks. 
% SigLit [] proposed a VLM training procedure where the image encoder remains fixed (locked) during training. It was demonstrated that this procedure is beneficial for developing a comprehensive VLM.
SigLip~\citep{zhai2023sigmoid} proposed a sigmoid loss function for language-image pre-training, improving alignment and robustness, particularly in handling diverse and noisy datasets. 
MaMMUT~\citep{kuo2023mammutsimplearchitecturejoint} presented a streamlined architecture with a vision encoder and text decoder, achieving state-of-the-art performance across diverse multi-modal tasks, including image-text retrieval, video question answering, and open-vocabulary detection.
SigLiT~\citep{zhai2022litzeroshottransferlockedimage} (Sigmoid Loss with Locked-Image Tuning) uses a pretrained Vision Transformer (ViT) from ImageNet-22K and fine-tunes only the text encoder on image-caption pairs. 
% It employs a pairwise sigmoid loss for efficient image-text pretraining, achieving strong zero-shot performance.
It employs a simplified loss function for efficient training, achieving strong zero-shot performance.
% ~\cite{minderer2022simpleopenvocabularyobjectdetection}. 

Several studies have been exploring the application of VLMs on the remote-sensing domain. RemoteCLIP~\citep{liu2023remoteclip} adapts CLIP for remote sensing, leveraging image-text pairs derived from existing remote-sensing datasets. SkyScript~\citep{wang2023skyscript} aligns OpenStreetMap features on remote sensing images to create a template-based RS image captioning dataset. Several works used LLM-modified RS-datasets to train RS-VLMs (e.g, LHRS bot by \citet{muhtar2024lhrsbotempoweringremotesensing}, GeoChat by \citet{kuckreja2023geochatgroundedlargevisionlanguage}, Geo-RSCLIP by \citet{Zhang_2024}). However, the advances in remote-sensing foundation models demonstrated by these studies remain constrained by the scale and semantic diversity of these datasets, leaving a large margin to improve generalized VLMs for remote sensing.

% Geo-RSCLIP \citep{Zhang_2024} use both LLM-modified RS datasets, together with RS-filtered image-text datasets, to create a RS caption dataset. Both works train a VLM on these datasets and present zero-shot results.
% \textbf{Remote Sensing VLMs}.
% The application of generalized VLMs in remote sensing is comparatively sparse. At first,
% several works introduce a remote-sensing focused chat-bot. An example is GeoChat \refp{kuckreja2023geochatgroundedlargevisionlanguage}, that used Vicuna LLM to construct a RS instruction tuning dataset from existing RS detection datasets, and use it to fine-tune a Vicuna based architecture. In addition, LHRS bot \citep{muhtar2024lhrsbotempoweringremotesensing} use the Open Street Map (OSM) data source to align map features on remote sensing images, and use LLM to convert to a RS image-language captioning dataset, in order to create RS-focused chat bot. Several works focused on zero-shot retrieval and classification for remote sensing images. An example is SkyScript \citep{wang2023skyscript}, which align OSM map features on remote sensing images to create a template-based RS image captioning dataset. In addition, Geo-RSCLIP \citep{Zhang_2024} use both LLM-modified RS datasets, together with RS-filtered image-text datasets, to create a RS caption dataset. Both works train a VLM on these datasets and present zero-shot results.

\section{Methodology}

% To address the gaps in remote sensing vision-language dataset, this section outlines two methods for generating unique remote sensing datasets, used to train a \Yehonathan{small} robust remote sensing 800M-parameters VLM, where its design potentially make it feasibale to run it efficiently on a resource-constrained device. 
This section introduces two methods for creating unique remote sensing vision-language datasets to address the absence of rich textual descriptions in remote sensing images and the scarcity of labeled data. 
These datasets are then used to train a compact yet robust 800M-parameter VLM for remote sensing, designed with the potential to run efficiently on resource-constrained devices.
We leverage the pre-trained foundation VLM, MaMMUT~\citep{kuo2023mammutsimplearchitecturejoint}, which is a model pre-trained with a contrastive learning approach.
To fine-tune it on our two novel datasets, namely RS-WebLI and RS-Landmarks, we use the shape-optimized $400$M-parameter ViT as a vision encoder~\citep{dosovitskiy2021imageworth16x16words}, while the language model has additional $400$M parameters.

\subsection{Remote-sensing dataset creation}

% In this work, we build upon an of-the-shelf VLM, i.e. a two-tower model trained in a contrastive manner with zero-shot capabilities. We fine-tune this model over our two novel datasets, RS-WebLI and Google Maps (both separately and together). Our goal is to .....
% Following we describe our two remote-sensing dataset-generation method. 
% To fill in the dataset gaps in remote sensing language data, in this section we describe our two methods to generate a unique remote-sensing datasets

\textbf{RS-Landmarks Dataset.} The RS-Landmarks dataset is a novel remote-sensing dataset comprised of $18$ million images with high-quality textual descriptions generated with Gemini $1.5$ Pro \citep{team2023gemini} as a teacher model. Initially, the satellite and aerial images are aligned with locations and footprints of places and landmarks extracted from Google Maps\footnote{Google Maps data used with permission from Google}, found within the associated images. We feed this information, alongside with the image, to the Gemini model, and use a tailored, curated prompt to generate concise captions for each image. This process yields high-quality and informative captions, describing a diverse set of object categories. 
 
\paragraph{RS-WebLI Dataset.} The RS-WebLI dataset is comprised of aerial and satellite imagery taken from the WebLI dataset~\citep{chen2023palijointlyscaledmultilinguallanguageimage}. We generated RS-WebLI by training aerial and overhead classifiers and thereafter using the classifiers to filter the WebLI dataset. First, we manually classified several hundred RS images in the WebLI dataset, using a simple caption heuristic and manual inspection of the dataset. Second, using the manually labeled sample, we trained an image classifier for remote sensing images and generated a dataset consisting of 40K images. Third, we harnessed the capabilities of crowd computing and initiated a labeling task wherein participants classified images as overhead aerial or satellite imagery, angled aerial imagery, or none of the above. The results were combined with random negatives, yielding a $60$K labeled dataset. 
Lastly, using the large dataset, we trained new aerial and overhead classifiers and applied them across the entirety of WebLI, filtering the data of which we chose $3$ million clean images, creating the RS-WebLI dataset.

\subsection{Training} 
Following \citet{zhai2023sigmoid}, \citet{kuo2023mammutsimplearchitecturejoint} and \citet{chen2022pali},
we start with the \underline{M}aMMU\underline{T} model, pre-trained for 500K steps on the WebLI dataset, denoted as the \textit{MT-WebLI} model. We then train two baseline models \textit {MT-RSWebLI} and \textit{MT-RSLandmarks} on the RS-WebLI and on RS-Landmarks datasets respectively, each for 20K steps. Finally, we train the combined \textit{MT-RSWebLI-RSLandmarks} model in an optimized tri-step curriculum: (i) for 20K steps over the RS-WebLI dataset, (ii) for additional 20K steps over the RS-Landmarks dataset, and (iii) for another 5K steps on a mix of the two. 
% Moreover, to show the benefit of each of our datasets separately, we fine-tuned our base on RS-WebLI and on RS-Landmarks alone, each for 20K steps, and refer to them as \textit {MT-RSWebLI} and \textit{MT-RSLandmarks} respectively. 

All training steps (including the pre-training) were done with a batch size of 16K, using a Sharded Adafactor optimizer with Adam learning rate decay. The learning rates used are 1e-3, 1e-6, 5e-6 and 1e-7, for the WebLI pre-training, RS-WebLI training, RS-Landmarks, and mix phases correspondingly. In the mix training phase, we use a mix of 97\% RS-Landmarks and 3\% of RS-WebLI.

\subsection{Zero-shot evaluations} 
 
%  \Yehonathan{Yotam - you guys must explain in Appendix how you conduct the training (hyper-parameters, epoch number, batches,....)}}

% Zero-shot Classification Evaluation method

We evaluated the zero-shot retrieval performance of our proposed model, as presented in Table~\ref{Tab1}. We used a nearest neighbor approach on the output embeddings of the model to match each image to a class. Following standard practice, we report the average top-1, top-5, and top-10 recall scores. If a dataset provided multiple captions for a single image, we considered the retrieval a success if any of the correct captions were selected. The table shows that MT-RSWebli-RSLandmarks outperforms all other public models. Moreover, using both datasets is essential to obtain optimal results.

\begin{table*}[h]
\centering
\caption{Average of top-1/5/10 of zero-shot retrieval results for image to text (I2T) and text to image (T2I). Our result are compared to PIR-ITR    \citet{pan2024pir}, SkyScript \citet{wang2023skyscript}, and \citet{zhang2023rs5m} (Geo-RSClip).
 Our base model was trained on the WebLI dataset.
 }
\begin{tabular}{lcccccccc}
\toprule
\multirow{2}{*}{\textbf{Model Name}} & \multicolumn{2}{c}{\textbf{RSICD}} & \multicolumn{2}{c}{\textbf{UCM Cap.}} & \multicolumn{2}{c}{\textbf{NWPU}} & \multicolumn{2}{c}{\textbf{RSITMD}} 
\\
& I2T & T2I & I2T & T2I & I2T & T2I & I2T & T2I \\
\midrule
PIR-ITR & 24.43 & 25.77 & - & - & - & - & 38.64 & 39.85 \\
SkyScript SkyCLIP-30 & 23.70 & 19.97 & 72.22 & 59.33 & - & - & 30.75 & 30.58 \\
Geo-RSClip+RS5M & 26.41 & 25.96 & - & - & - & - & 33.33 & 38.02 \\
\midrule
MT-WebLI                & 23.88 & 24.17 & 69.21 & 66.50 & 20.33 & 23.18 & 28.83 & 32.70 \\
MT-RSWebLI             & 26.68 & 25.09 & 72.38 & 65.96 & 24.62 & 22.60 & 32.15 & 35.57 \\
MT-RSLandmarks         & 33.12 & 32.98 & 72.38 & 70.91 & 40.10 & 32.15 & 41.96 & 42.30 \\
MT-RSWebli-RSLandmarks & \textbf{33.33} & \textbf{33.59} & \textbf{74.76} & \textbf{71.79} & \textbf{41.44} & \textbf{32.28} & \textbf{42.63} & \textbf{42.58} \\
\bottomrule
\end{tabular}
\label{Tab1}
\end{table*}

%\Yotam{In addition, in order to evaluate the model generalization outside of the classes that appear in the RS-Landmarks dataset, we perform an experiment in which we manually chose 89 landmark types, and remove every image that includes them from the dataset. We refer to this dataset as \textbf{RS-LandMark-holdout}. From the dropped images, we construct a new classification dataset, in which every image is centered around a landmark from the $89$ types above. We evaluate our model ability to recognize the image class using a simple nearest neighbhor technique. We compare the result of our Base model, trained solely on the WebLI dataset, and a model that was trained on RS-LandMark-holdout. The significant improvement that is shown in Table \ref{holdout} examplifies our model ability to generalize and perform well on unseen classes, beyond those in Google Maps.}

We demonstrate the benefit of utilizing our remote sensing datasets to train a model which is capable of generalizing on the remote-sensing domain. This result is further substantiated by showing that the enhanced accuracy over the non-RS baseline is kept even when testing on categories which \textit{were not} explicitly presented during training. To illustrate this, we created \textit{RS-Landmarks-89}, an image classification benchmark dataset, comprising 89 manually chosen landmark categories from the RS-Landmarks dataset, where each image is centred on one of the landmark types. Complementarily, the \textit{RS-Landmarks-89-holdout} image-text dataset was created by removing images from the RS-Landmarks dataset that contained the previously mentioned categories. To evaluate the zero-shot classification abilities on unseen categories, we fine-tune the MT-WebLI model with the \textit{RS-Landmarks-89-holdout} dataset. We evaluate the resulting model on the RS-Landmarks-89 classification dataset using simple nearest-neighbor approach. 

Unsurprisingly, Table 2 shows that a model tuned on the RS-Landmarks dataset outperforms the baseline, general purpose MT-WebLI model on the RS-LandMarks-89 dataset. 
The ability to generalize on the remote sensing domain is evaluated by the model tuned on the RSLandMarks-89-holdout dataset (2nd row). The performance of this model, which was not trained on any of the 89 categories, is significantly higher than the MT-WebLI baseline, and is close to that of the full MT-RSLandmarks.

% The accuracy of this model, along with the MT-WebLI and MT-RSLandmarks models, are presented in Table \ref{holdout}.
% Our tuned models exhibit notably higher accuracy (Table \ref{holdout}, rows 2-3) than the baseline (Table \ref{holdout}, 1st row) on the \textit{RS-Landmarks-89} classification task, achieved by training on remote-sensing image-captions from our \textit{RSLandmarks} dataset. This improvement is largely maintained even without training on the landmark types in RS-Landmarks-89-holdout (Table \ref{holdout}, 2nd row), showing our VLM is capable of generalizing on the remote-sensing domain.

\begin{table*}[h]
\centering
\caption{\textbf{Generalization to unseen classes}. Nearest neighbour classification accuracy on the holdout dataset. The 89 classes in the \textit{RS-Landmarks-89} are excluded from the \textit{RS-Landmarks-89-holdout} training-set.}
\begin{tabular}{lccccc}
\toprule
\textbf{Model Name} & \textbf{RS-Landmarks-89} \\
\midrule
MT-WebLI & 13.78 \\
MT-WebLI tuned on RSLandmarks-89-holdout & 21.38 \\
MT-WebLI tuned on RSLandmarks &  25.42 \\
\bottomrule
\end{tabular}
\label{holdout}
\end{table*}

\section{Self-supervised zero-shot localization via iterative refinement} 
% \Yehonathan{TODO - Ill merge the sigmentation overleaf here + using aviad new figures}

\begin{figure}[h]
\centering
    \centering
    \includegraphics[width=0.99\linewidth]{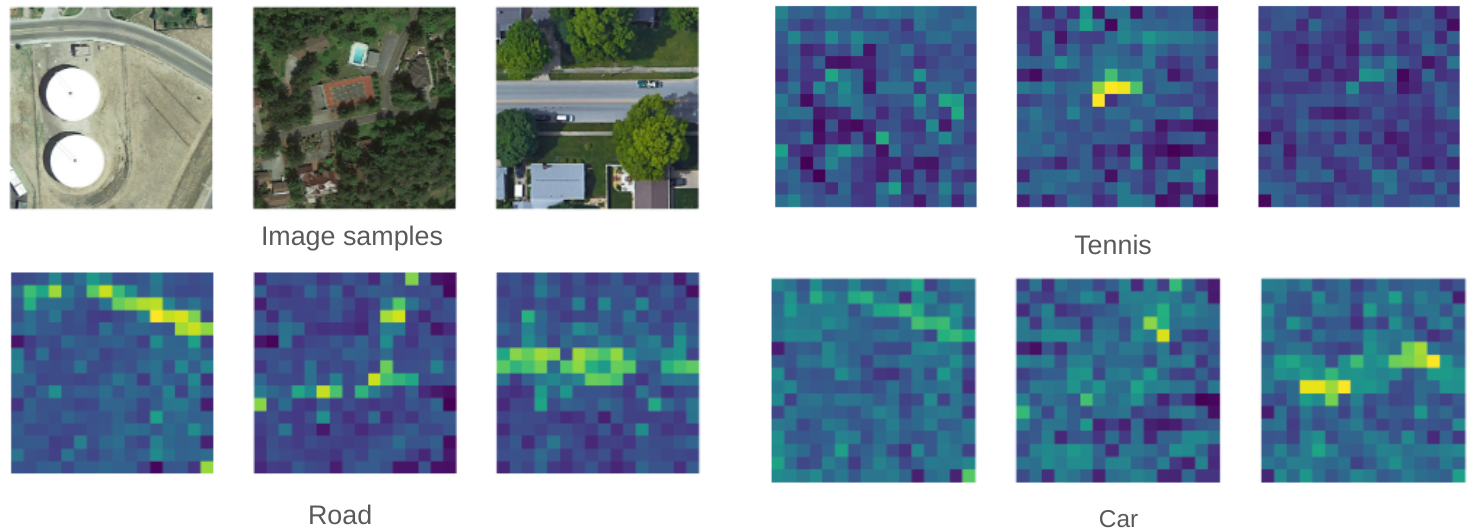}
\caption{Attention similarity maps for text queries (\textit{'Tennis', 'Road', 'Car'}) in different images, selected from the DIOR dataset~\citep{dior2023rs}. }
\label{fig:combined} % Label for the combined figure
\end{figure}

% Our approach distills implicit model knowledge, as an image proxies, to enhance few-shot and zero-shot learning on novel classes. This is done using the following surprising insight. Despite being explicitly trained for multi-class classification, the model appear to perform classification at the segment patch-size level, on the attention-polling layer, which is illustrated in Figure \ref{fig:combined}\Yehonathan{todo - figure + get better pic form Aviad}, where the image patch level values are corresponeded to the similarity between the novel given class and the patch information. \Yehonathan{explain local-global information phenomena}                      
% In this section, we present the ongoing effort to distill implicit knowledge from such VLM model for pseudo-labeling by leveraging image patches as proxies to enhance few-shot and zero-shot learning.
In this section, we present the ongoing effort to distill image-level knowledge gained in the VLM contrastive training procedure to enhance the model's localization ability. Enhanced localization is essential for performance in downstream vision tasks, notably segmentation. Applying the methodology described in the previous section, with an additional attention pooling head similarly to SigLIP\footnote{This model has the SoViT-400m architecture, which is the shape-optimized version as presented in ~\cite{alabdulmohsin2024gettingvitshapescaling}}, reveals an important insight, that despite being explicitly trained for image-level on broad captions, the model appears to classify individual patches at the attention-pooling layer. This phenomenon is illustrated in Figure~\ref{fig:combined}, where the similarity is captured at the patch-level.

% Using the methodology described in the previous section but this time to train a LLM with attention pooling head reveals the surprising insight: despite being explicitly trained for multi-class classification, the model appears to perform classification at the segment (patch-size) level in the attention-pooling layer. This phenomenon is illustrated in Figure~\ref{fig:combined}, where the similarity between the novel class and the image patch-level information is captured. In this section we present the ongoing effort to distills implicit knowledge from such VLM model (including attention pooling head), potentially SigLIP\footnote{This model has the SoViT-400m architecture, which is the shape-optimized version as presented in \cite{alabdulmohsin2024gettingvitshapescaling}}, for pseudo-labeling by leveraging image patches as proxies to enhance few-shot and zero-shot learning on novel classes.

% In this section we present the ongoing effort to distills implicit knowledge from our VLM for pseudo-labeling by leveraging image patches as proxies to enhance few-shot and zero-shot learning on novel classes. This is based on a surprising insight: despite being explicitly trained for multi-class classification, the model appears to perform classification at the segment (patch-size) level in the attention-pooling layer. This phenomenon is illustrated in Figure~\ref{fig:combined}, where the similarity between the novel class and the image patch-level information is captured.

We build upon this phenomenon to present a pseudo labeling algorithm that yields segmentation masks for a query of interest. Our algorithm begins with the model receiving the text query along with a stack of images. The query is paired with the images during inference, where for each image we create a similarity-based attention map (Figure~\ref{fig:combined}). This map segments the associated patches that match the text query within every image. 
A pre-defined adaptive similarity threshold is then applied to identify patches whose similarity to the query exceeds the threshold.  
We propose to use these pseudo-labeled samples to fine-tune the model.
This self-supervised process is repeated iteratively, with the adaptive similarity threshold gradually increasing as the model achieves higher accuracy in producing robust segmentation masks for arbitrary text queries.

To mitigate the noise in the segmentation maps, we introduce a novel approach called \emph{Smooth-Attention-Operation}. 
This method differs from standard attention mechanisms by applying the attention pooling layer, not to the entire input key, but rather to smaller, sliding window blocks of the key. The sliding window traverses the image, pooling attention values within its boundaries.  
Smaller windows capture highly localized attention, while larger windows incorporate more contextual information from neighboring regions.  
This approach balances local detail with broader context, offering a trade-off between per-patch similarity and full-image attention.

\section{Discussion}
In this work, we addressed the significant challenges of applying Vision-Language Models (VLMs) in the remote sensing domain. 
To overcome the limitations posed by the scarcity of labeled image-text data and the unique characteristics of remote sensing imagery, we introduced two novel datasets: RS-WebLI, a large-scale dataset derived from filtering and curating aerial and satellite imagery, and RS-Landmarks, a dataset enriched with high-quality captions generated by the Gemini teacher model using information extracted from Google Maps. 
These datasets provide a foundation for training VLMs specifically tailored to the remote sensing domain, enhancing their generalization capabilities. Leveraging these datasets, we trained a VLM foundation model that demonstrated state-of-the-art cross-modal retrieval performance across public benchmarks. Finally, we present a self-supervised zero-shot retrieval fine-tuning scheme for pseudo-labeling. 
% Our scheme is based on the surprising insight we witnessed that our model, despite being explicitly trained for multi-class classification, performs classification at the segment level in the attention-pooling layer.
Our scheme is based on the important observation that our model, despite being trained at image level similarity on broad captions, is able to perform remote sensing classification at the segment level in the attention-pooling layer.

\bibliographystyle{iclr2025_conference}
\bibliography{iclr2025_conference}

\clearpage
\appendix
\section{Appendix}
\subsection{Zero-shot classification evaluation}
We evaluate the model's zero-shot classification performance on remote sensing images without prior training on those specific classes using several remote sensing image classification datasets, presented in Table \ref{T2}. 
For each dataset, we created a set of descriptive sentences in the format "An aerial image of $<$class name$>$." 
Then, using a nearest neighbor algorithm, we determined the best matching class for each image based on these sentences.

% We could also mention the opposite case where there is more than one image per caption, but the handling is a bit trivial.

\begin{table*}[h]\label{T2}
\centering
\caption{Comparison of the Top-1 accuracy zero-shot classification performance, results taken from papers.}
\begin{tabular}{lccccc}
\toprule
\textbf{Model Name} & \textbf{FMOW} & \textbf{SkyScript} & \textbf{RESISC45} & \textbf{UCM Class.} & \textbf{AID} \\
\midrule
SkyScript & 28.04 & (70.89) & 70.94 & - \\
RS-CLIP & -     & 68.84 & 71.35 & 74.28 / 78.00 & 70.51 \\
GeoRSCLIP-VitL  & -     & -     & 71.89 & -     & 76.33 \\
GeoRSCLIP-VitH  & -     & -     & 73.83 & -     & 73.72 \\
RemoteCLIP      & -     & -     & 79.84 & -     & 91.3 \\
GeoChat (7B)    & -     & -     & -     & 84.43 & 72.03 \\
LHRS-Bot (7B)   & -     & -     & -     & -     & 91.26 \\
\midrule
MT-WebLI & 37.58 & 58.66 & 66.93 & 76.52 & 71.46 \\
MT-RSWebLI & 42.73 & 65.16 & 70.91 & 83.52 & 75.78 \\
MT-RSLandmarks & 42.93 & 66.31 & 68.55 & 77.67 & 73.15 \\
MT-RSWebli+RSLandmarks & 47.24 & 69.46 & 72.31 & 80.29 &  71.96 \\
\bottomrule
\end{tabular}
\end{table*}

\end{document}

%% file: math_commands.tex
\usepackage{amsmath,amsfonts,bm}

\def\eqref#1{equation~\ref{#1}}
\def\1{\bm{1}}

\DeclareMathAlphabet{\mathsfit}{\encodingdefault}{\sfdefault}{m}{sl}
\SetMathAlphabet{\mathsfit}{bold}{\encodingdefault}{\sfdefault}{bx}{n}